%% file: paper.tex
\pdfoutput=1

\documentclass[11pt]{article}

\usepackage[]{acl}
\usepackage{graphicx}

\usepackage{times}
\usepackage{latexsym}
\usepackage{array,multirow}
\usepackage{booktabs}
\usepackage{caption}
\usepackage{subcaption}
\usepackage{xcolor}
\usepackage[T1]{fontenc}
\usepackage{fancyhdr}

\usepackage[utf8]{inputenc}

\usepackage{todonotes}

\usepackage{microtype}

\definecolor{light-gray}{gray}{0.95}
\newcommand{\code}[1]{\colorbox{light-gray}{\texttt{#1}}}
\newcommand{\atomic}{{ATOMIC$^{20}_{20}$ }}
\newcommand{\token}{T{\small O}K{\small EN} }
\newcommand{\tokenwospace}{T{\small O}K{\small EN}}

\title{\tokenwospace: Task Decomposition and Knowledge Infusion \\ for Few-Shot Hate Speech Detection}

\author{Badr AlKhamissi\thanks{$\;\;$Equal Contribution} \And Faisal Ladhak\footnotemark[1] \AND Srini Iyer \And Ves Stoyanov \And Zornitsa Kozareva \And Xian Li \And Pascale Fung \AND Lambert Mathias \And Asli Celikyilmaz \\ \\ Meta AI \And Mona Diab}

\begin{document}

\maketitle
\begin{abstract}
    
    Hate speech detection is complex; it relies on commonsense reasoning, knowledge of stereotypes, and an understanding of social nuance that differs from one culture to the next. It is also difficult to collect a large-scale hate speech annotated dataset. In this work, we frame this problem as a few-shot learning task, and show significant gains with decomposing the task into its "constituent" parts. In addition, we see that infusing knowledge from reasoning datasets (e.g. \atomic) improves the performance even further. Moreover, we observe that the trained models generalize to out-of-distribution datasets, showing the superiority of task decomposition and knowledge infusion compared to previously used methods. Concretely, our method outperforms the baseline by 17.83\% absolute gain in the 16-shot case.

\end{abstract}

\section{Introduction}

\textbf{Disclaimer}: {\color{red} \textit{Due to the nature of this work, some examples contain offensive text and hate speech. This does not reflect authors’ values, however our aim is to help detect and prevent the spread of such harmful content.}}\\

The task of automatically detecting \textit{Hate Speech} (HS) is becoming increasingly important given the rapid growth of social media platforms, and the severe social harms associated with the spread of hateful content. However, building good systems for automated HS detection is challenging due to the complex nature of the task. It requires the system to understand social nuance, such as which groups are being targeted by the hateful content. Prior work has shown that even humans cannot achieve a high agreement on whether or not a social media post constitutes HS \cite{rahman2021information}.

In this work, we explore whether decomposing HS detection into subtasks that correspond to the definitional criteria of what constitutes HS (i.e. the offensiveness of a post and whether it targets a group or an individual) \cite{Davidson2017AutomatedHS, Mollas2020ETHOSAO} would lead to systems that are more accurate and robust. In particular, we show that task decomposition leads to more sample-efficient systems for HS detection, by showing improved results in the few-shot setting. Moreover, we demonstrate that infusing commonsense knowledge by fine-tuning on the \atomic \cite{Hwang2021COMETATOMIC2O} and \textit{StereoSet} \cite{Nadeem2021StereoSetMS} datasets improve performance even further. Specifically, we observe an absolute improvement of $17.83\%$ in the 16-shot case over the baseline model ($\S$\ref{sec:knowledge_infusion}). Further, we show that the resulting models are more robust, and are able to achieve better performance than baseline methods in out-of-distribution settings ($\S$\ref{sec:ood_results}).  

\begin{figure}
    \centering
    \includegraphics[width=1\linewidth]{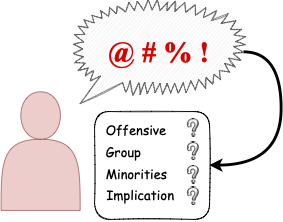}
    \caption{\textit{Hate Speech} decomposed into several subtasks leading to better results.}
    \label{fig:simpsons}
\end{figure}

Explainability is an important aspect for being able to identify and fix failure modes of HS systems \cite{attanasio-etal-2022-benchmarking}. 
To that end, we show that task decomposition of HS detection moves us a step closer towards explainable systems, allowing us to identify the problematic subtasks that may be the bottleneck for improving overall performance for HS detection. We then show that explicitly targeting to improve the problematic subtask leads to improved overall performance ($\S$\ref{sec:perf_analysis}).  

\begin{figure*}[ht]
\centering
\includegraphics[width=1\linewidth]{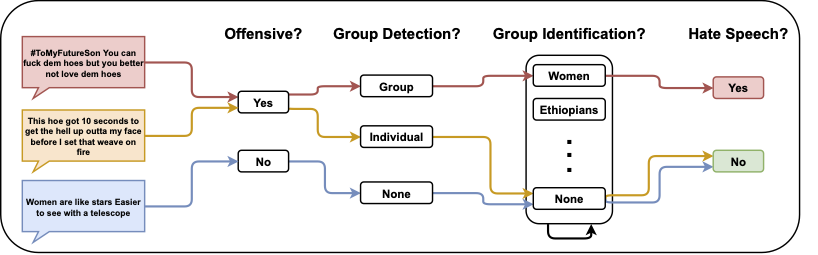}
\centering
\caption{Examples from the SBIC dataset. The post is classified as HS if it is \textit{Offensive} and a \textit{Group} is referenced as the target of the offensive speech.}
\label{fig:sbic}
\end{figure*}

The remainder of the paper is structured along four axes. We first present the importance of (1) \textbf{Task Decomposition} and (2) \textbf{Knowledge Infusion} for training better few-shot HS detection models. Specifically, we compare our method against a baseline that only outputs a binary prediction for HS. Thus showing the significance of decomposing the prediction into several subtasks and pre-finetuning the model on two different reasoning datasets that instill in the model a degree of commonsense reasoning and knowledge of stereotypes. We evaluate our experiments across 10 seeds each of which uses a different sampled dataset with sizes ranging from 16 to 1024 samples. Our method, \tokenwospace, show significant improvements\footnote{We use Welch's $t$-test on all experiments that were repeated 10 times, and consider $p<0.05$ as significant.} over the baseline. The trained models can also (3) \textbf{Generalize} better to three out-of-distribution datasets, and are more (4) \textbf{Robust} to training data and hyperparameters. We demonstrate this by measuring the variance across different seeds, data partitions and hyperparameters and show that it is significantly smaller for the \token models compared to the baseline.  





\section{Few-Shot Hate Speech Detection}
Collecting a high quality large-scale dataset for HS is difficult since it is a relatively rare phenomenon, which makes it hard to sample social media posts containing HS without relying on keywords that may be indicative of it \cite{rahman2021information}. However, despite being rare, its effects are of significant harm. Further, since HS is a complex phenomenon, relying on keywords may result in datasets with low coverage that are not effective in capturing more subtle forms of HS \cite{elsherief-etal-2021-latent}. This further results in building models that are less generalizable and can exhibit racial biases \cite{davidson-etal-2019-racial, sap-etal-2019-risk}.

Motivated by these, we frame HS detection as a few-shot learning task, where the model is given a limited number of examples to learn what constitutes HS, and explore whether we can build robust HS detection models that can generalize well in cases where we do not have a lot of training data. In particular, we show that our trained \token models are more \textit{generalizable} by measuring the performance on out-of-distribution HS datasets, and are more \textit{robust} by measuring the variance in performance of HS detection across different randomly sampled few-shot datasets and hyperparameters. 

\section{Datasets} 


\paragraph{SBIC \cite{sap2020socialbiasframes}.} 
We use the Social Bias Inference Corpus (SBIC) to construct few-shot training and validation sets. This corpus includes posts from several online social media platforms, such as Reddit, Twitter, etc., along with the annotations for the offensiveness, targeted group as well as the implications to further explain what stereotype is being implied by the post.
While the dataset does not have explicit labels for whether or not a post is HS, we derive it using the annotations for offensiveness and group detection, i.e. a post is considered HS if it contains offensive/derogatory language that is expressed towards a targeted group (see Figure \ref{fig:sbic}). This is consistent with the definition of HS used by prior work \cite{Davidson2017AutomatedHS, Mollas2020ETHOSAO}.

To construct few-shot training sets, we perform a stratified sampling of data from the SBIC corpus, up to a target size $n$. We sample $\frac{n}{4}$ examples containing inoffensive posts, $\frac{n}{4}$ examples containing offensive, but non-HS posts. The remaining budget of $\frac{n}{2}$ samples is used for posts containing HS, spread evenly across different targeted groups to ensure diversity. We create datasets of varying sizes from $16$ samples up to $1024$, ensuring each smaller dataset is a proper subset of the larger datasets. We sample $10$ different datasets for each target size $n$ using $10$ different random seeds.


\subsection{Reasoning Datasets}

\paragraph{\atomic \cite{Hwang2021COMETATOMIC2O}} 
This is a commonsense knowledge graph containing $1.33$M inferential knowledge tuples in textual format that are not readily available in pretrained LMs. \atomic encodes different social and physical aspects of the everyday experience of human life. In this work, we find that training on human readable templates in-place of each tuple vastly improves the downstream SBIC few-shot HS performance. Examples of such templates are shown in the Appendix ($\S$\ref{app:atomic_templates}).

\paragraph{StereoSet \cite{Nadeem2021StereoSetMS}} 
This dataset was developed to measure stereotype bias in LMs. Specifically, it contains $17$k sentences that measure biases across four different domains: gender, profession, race and religion. In this work, we only finetune task-decomposed models on a subset of StereoSet. In particular we only use stereotypes that belong to the intersentence task since we found that it results in better HS detection models. More details on the StereoSet training can be found in the Appendix ($\S$\ref{app:stereoset}).

\subsection{Out-of-Distribution Datasets}

In addition, to test the generalizability of our models, we use the following three corpora to evaluate out-of-distribution performance: 

\paragraph{HateXplain \cite{mathew2020hatexplain}.}  This corpus includes posts from Twitter and Gab along with the HS labels (i.e. hate, offensive or normal), target community (i.e. victim of the HS or offensive speech) and the rationales (i.e. spans from the posts that affected the annotator's decision). In our work, we convert the HS labels into a binary HS/non-HS label, to measure performance for HS detection.

\paragraph{HS18 \cite{gibert2018hate}.} This corpus consists of sentences from posts on Stormfront, a white supremacist forum, along with labels for HS.

\paragraph{Ethos \cite{mollas2020ethos}.} This corpus (compiled by researchers at the Aristotle University of Thessaloniki) consists of comments from social media platforms (YouTube and Reddit) with binary labels for HS, as well as a finer-grained categorization of the type of HS. We use the binary labels in our work to measure the performance for HS detection. Number of examples in test set of each evaluation dataset is shown in Table \ref{tab:data_stats}.

\begin{table}[ht]
\centering
\begin{tabular}{lc}
\toprule
\textbf{Dataset} & \textbf{\# Examples}  \\ \hline
HateXplain & 1,924\\ 
HS18 & 9,916\\ 
Ethos & 998\\ 
\bottomrule
\end{tabular}
\caption{Evaluation dataset statistics.}
\label{tab:data_stats}
\end{table}

\begin{table*}
\centering

\begin{tabular}{@{}llp{0.5\linewidth}@{}}
\toprule
                                      & \multicolumn{1}{c}{\textbf{Input}} & \multicolumn{1}{c}{\textbf{Output}}                                                                \\ \midrule
\textbf{Baseline}                     & Post: \{\texttt{POST}\} Hate speech?       & \{\texttt{HS}\}                                                                                             \\
\textbf{\token} & Post: \{\texttt{POST}\} Offensive?         & \{\texttt{OFF}\} Target implication? \{\texttt{GD}\} Targeted minorities? \{\texttt{GI$_1$, ..., GI$_\textsc{N}$}\} Hate speech? \{\texttt{HS}\} \\ \bottomrule
\end{tabular}

\caption{Linearization scheme for the \textbf{Baseline} and the \textbf{\token} models. Given the post, the \textit{Baseline} predicts whether it is HS or not; whereas the task-decomposed model does the prediction for \textit{Offensiveness}, \textit{Group Detection} and \textit{Group Identification} before predicting the HS label. \texttt{HS} and \texttt{OFF} are binary labels (i.e. either \code{Yes} or \code{No}). \texttt{GD} can be one of \{\code{Group}, \code{Individual}, \code{None}\}. Finally, \texttt{GI$_{\textsc{i}}$} is a group identity (e.g. \code{Women}).}
\label{tab:linear_scheme}
\end{table*}

\section{Experimental Setup}

\subsection{Task Decomposition}
\label{sec:task_decomposed_methods}

HS detection is a complex and subjective task, and prior work has shown that it is hard to get high agreements between humans about whether or not a post constitutes HS \cite{sanguinetti-etal-2018-italian, assimakopoulos-etal-2020-annotating}. Therefore recent efforts on hate speech annotation have turned to more fine-grained, hierarchical annotation schemes that break HS detection into subtasks that correspond to the definitional criteria of what constitutes HS, leading to higher agreement scores than reported by prior work \cite{assimakopoulos-etal-2020-annotating, mathew2020hatexplain, sap2020socialbiasframes, rahman2021information}. 

Motivated by these findings, we treat HS detection as a conditional generation task, since that allows us to represent classification and generation subtasks in a unified framework. The model is given the set of tokens in a post as input and is tasked with generating the inferences related to the HS subtasks. Table \ref{tab:linear_scheme} shows the linearization scheme that we use to train our models. The baseline model is tasked with generating a binary prediction for HS, whereas the task-decomposed model has to generate the predictions for the subtasks, in order, before making the prediction for HS. Specifically, the model predicts first if the post is offensive, then whether it is targeting a group or an individual or neither, and following that it predicts the identities of the targeted groups (e.g. disabled people). This forces the task decomposed model to reason about the subtasks before deciding whether or not a post constitutes HS. For all sets of experiments, we finetune the pre-trained BART\textsubscript{LARGE} model \cite{lewis-etal-2020-bart} provided by the HuggingFace library \cite{Wolf2019HuggingFace} for the task of HS detection. The results are shown in Table \ref{tab:task_decomposed_results}. 

\input{tables/task_decomposition}

\subsection{Knowledge Infusion}

In the following set of experiments we ask whether incorporating commonsense knowledge and stereotypes into the model show significant improvements on the few-shot HS detection task. 

\paragraph{\atomic}
To answer this question, we first finetune BART on the \atomic dataset, where each tuple is converted into a natural language statement using human readable templates (see $\S$\ref{app:atomic_templates}). The resulting model achieves similar performance on the held-out \atomic test set as the one reported in \citet{Hwang2021COMETATOMIC2O}. Following that, we further finetune the resulting model on both the baseline and the task decomposed data from SBIC as described in $\S$\ref{sec:task_decomposed_methods}. Results are shown in Table \ref{tab:knowledge_infusion_results}.  

\paragraph{StereoSet}
Here, we finetune both BART and our model finetuned on \atomic on stereotypical sentences from the StereoSet dataset. Specifically, we similarly treat it as a conditional generation task, where we predict the context based on the sentence, bias type and target group (see Appendix $\S$\ref{app:stereoset}).
 
\section{Results}
\label{sec:results}

\begin{figure*}[ht]
    \centering
    \begin{subfigure}{0.33\linewidth}
        \includegraphics[width=1\linewidth]{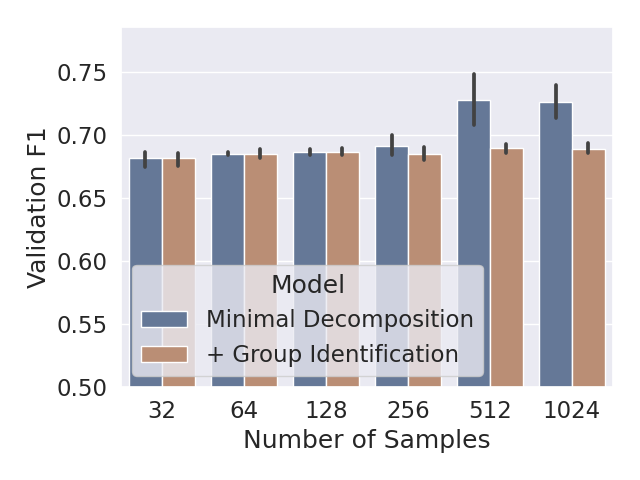}
        \caption{Offensiveness Performance}
        \label{fig:offensive_performance}
    \end{subfigure}%
    \begin{subfigure}{0.33\linewidth}
        \includegraphics[width=1\linewidth]{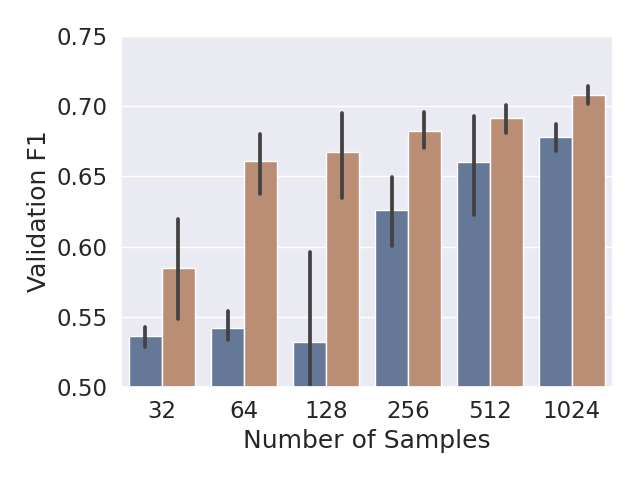}
        \caption{Group Detection Performance}
        \label{fig:groupd_performance}
    \end{subfigure}%
    \begin{subfigure}{0.33\linewidth}
        \includegraphics[width=1\linewidth]{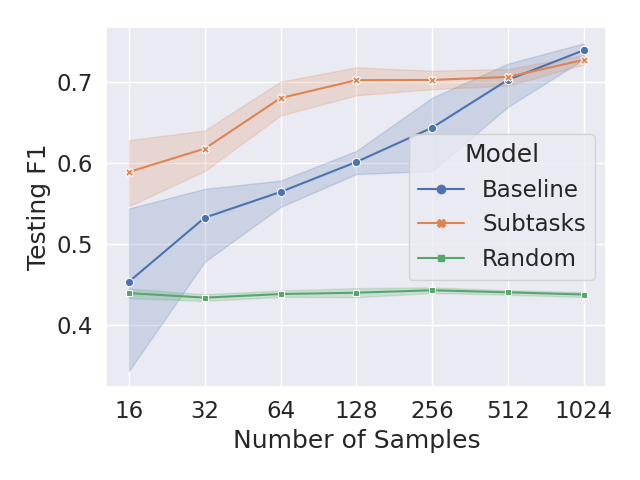}
        \caption{Hate Speech Performance}
         \label{fig:bart_comparison}
    \end{subfigure}
    \caption{\textbf{(a)} The validation F1-score of the \textit{Offensiveness} subtask for the minimal decomposition and task-decomposed models. \textbf{(b)} Similarly, this is the validation F1-score for the \textit{Group Detection} subtask. It can be seen that adding the \textit{Group Identification} subtask improves the performance dramatically. \textbf{(c)} Testing Performance of \textit{Baseline} vs the \textit{Task Decomposed} model. Random performance plotted for reference.}
\end{figure*}

In this section, we report the mean binary F1-scores on the testing set across $10$ different seeds for each dataset size for the HS detection task. Specifically, we compare the \textbf{Baseline} model, which directly predicts a binary label of whether the post is HS or not, with the \textbf{\token} model that employ task decomposition and knowledge infusion as described in the previous section.

\subsection{Task Decomposition} 

Table \ref{tab:task_decomposed_results} shows the effect of task decomposition on the HS detection task. In particular, we compare the baseline model with a model that uses only the \textit{Offensiveness} and \textit{Group Detection} tasks as it's subtasks before predicting the HS label. This is referred to as the \textbf{Minimal Decomposition} model since it uses the minimal constituents that we used to derive the HS label. Following that, we add the \textbf{Group Identification} to the subtasks and observe significant improvements over the \textit{Baseline} in the few-shot setting. However, as the dataset size increases beyond $512$ samples, the observed differences in the mean are no longer significant (see Figure \ref{fig:bart_comparison}). 

\subsubsection{Fine-Grained Error Analysis} 
\label{sec:perf_analysis}

Task decomposition allows us to perform finer grained error analysis to identify failure modes of the HS model. Specifically, we analyze whether \textit{Offensiveness} classification or \textit{Group Detection} is more challenging for the model to learn. Figure \ref{fig:offensive_performance} shows the performance of the model for the \textit{Offensiveness} subtask, while Figure \ref{fig:groupd_performance} shows the performance for \textit{Group Detection} subtask with varying number of training samples. 

We observe that the overall performance for \textit{Group Detection} subtask consistently lags behind \textit{Offensiveness} prediction, especially when we have fewer examples in the training data. We note that the model is able to achieve a reasonable performance ($\sim68\%$) for \textit{Offensiveness} prediction even in the few-shot regime. This suggests that \textit{Group Detection} subtask is the bottleneck in improving the performance for HS classification, and in order to improve further we need our model to be more accurate for this subtask.

Given these findings, we further explore whether adding more fine-grained information related to groups would help improve the \textit{Group Detection} subtask. In particular, we additionally require the task-decomposed model to generate the group that is being targeted by the offensive content in the post. Figure \ref{fig:groupd_performance} shows the performance of the model in predicting \textit{Group Detection} when we require the subtask model to identify the groups that are being targeted. We see that incorporating this subtask significantly improves the performance for \textit{Group Detection} in the few-shot setting. Figure \ref{fig:bart_comparison} shows that this further translates to improved HS detection in the same regime. 


\input{tables/knowledge_infusion}

\subsection{Knowledge Infusion} 
\label{sec:knowledge_infusion}

Table \ref{tab:knowledge_infusion_results} show the binary F1-scores across 10 seeds on the SBIC testing-set for the HS detection task using different knowledge infused models for both the baseline and task-decomposed datasets (referred as Subtasks). The first row show the BART model without any knowledge infusion (same as the one reported in Table \ref{tab:task_decomposed_results}). The following row show the results when we finetune the pre-trained BART on StereoSet. It can be seen that this result is the best performance in the 32-shot regime. In the third row we finetune BART on the natural language version of the \atomic dataset. This increases performance most noticeably in the 16-shot regime. The final row shows the results when we further finetune the model finetuned on \atomic on StereoSet, it consistently improves the performance even further from the 64-shot setting to the 512-shot setting. In all models, the difference between the baseline and subtasks model is significant in most cases until we reach 512 training examples. It can be seen that knowledge infusion alone does not seem to consistently improve performance over the BART baseline model, however when combined with subtask decomposition, it leads to the best results overall. This would imply that the reasoning knowledge helps the model to better understand relationships among subtasks.

\subsection{Generalizability} 
\label{sec:ood_results}

\begin{figure*}[ht]
    \centering
    \includegraphics[width=1\linewidth]{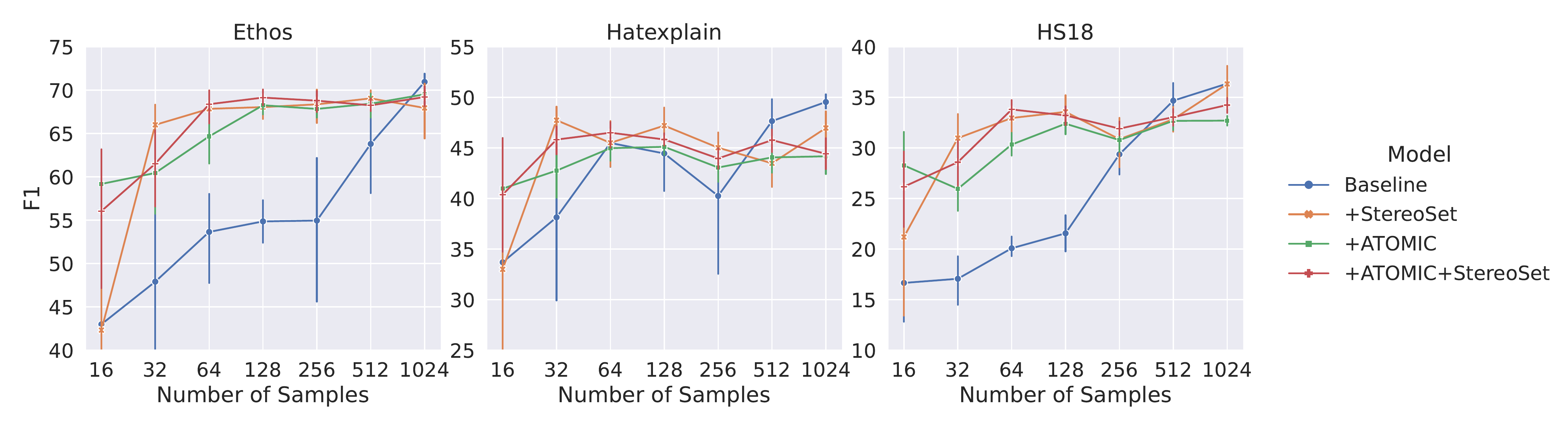}
    \caption{\textbf{OOD Results} Performance of Baseline versus \token models that employ different degrees of knowledge infusion on OOD datasets HS18, Ethos, Hatexplain.}
    \label{fig:zeroshot_OOD}
\end{figure*}

Figure \ref{fig:zeroshot_OOD} compares the OOD performance of the baseline model with subtasks models that were trained with different degrees of knowledge infusion. Note that all models were trained on the SBIC data and evaluated for each of the three datasets in a zero-shot manner, i.e. we do not perform any further dataset specific finetuning. Similar to above, each point represents the mean F1 scores across $10$ different runs for the given dataset size. We see that for all three datasets, the \token models consistently outperforms the baseline, and shows significantly better zero-shot generalizability in the few-shot setting. However, we note though that HateXplain has a slight distribution shift (since it's also built based on Twitter as SBIC), and interestingly it can be seen that only the results from Ethos and HS18 (in Figure \ref{fig:zeroshot_OOD}) are significantly different from the binary (baseline) prediction model. This implies that our model is indeed better at generalizing to out-of-distribution data.

\subsection{Robustness}
Here, we show that the \token models are less sensitive to changes in hyperparameters and training dataset selection. Specifically, Table \ref{tab:robustness} show the average standard deviation across different hyperparameters on the validation set as a function of the number of training samples. It can be seen that the subtasks model is significantly more robust than the baseline model. 

\begin{table}
\centering
\begin{tabular}{@{}ccc@{}}
\toprule
\textbf{\# of Samples} & \textbf{Baseline} & \textbf{\token} \\ \midrule
16                     & 15.75             & 2.44              \\
32                     & 14.15             & 0.55              \\
64                     & 13.71             & 0.87              \\
128                    & 16.44             & 1.66              \\
256                    & 21.20             & 2.69              \\
512                    & 17.29             & 3.36              \\
1024                   & 15.72             & 3.20              \\ \bottomrule
\end{tabular}
\caption{\textbf{Robustness Results} Here we report the average standard deviation across the 10 runs for a given training set size across the different hyperparameters used in our experiments. It can be seen that the task-decomposed model is more robust to training configuration.}
\label{tab:robustness}
\end{table}

\section{Analysis}

In this section, we explore whether the order of the subtasks matter or not, the scale of the model and if adding an additional subtask that requires the model to generate natural language explanations as to why the post is considered HS helps improve the overall HS performance. The results reported here were done across only 5 seeds. In addition, we show a few examples that the baseline model was not able to capture but the \token model got correctly.

\subsection{Does Order Matter?}

Table \ref{tab:hs_pos_results} shows the importance of the order of the HS label in the sequence of subtasks. It can be seen that placing HS at the end gets the best result when the number of training samples are $64$ or less. 

\input{tables/hate_speech_position}

\subsection{Do Implications Matter?}

Here, we add an additional subtask with the goal of generating an implied stereotype in natural language for a HS post or `None' otherwise. For example, given the following post: ``\textit{How do you make a Muslim's phone explode? Set it to airplane mode.}'' The implied stereotype would be: \code{Muslims are terrorists.} Therefore, the model is tasked with generating such a sentence before predicting the HS label. Table \ref{tab:implications} shows the HS detection performance before and after adding the implications subtask in comparison with the baseline across 5 seeds for a given training set size. It can be seen that although adding the implied stereotype to the list of subtasks pushes the model to performing better than the baseline when the number of samples is less than 512, it still falls short to the \token model without the implication. The reason behind this might be because the implications were noisy and sometimes too generic, which is why it might have resulted in performance degradation. Further, we believe that scaling up the model to be an order or two magnitude larger will enable a better utilization of the implications.

\input{tables/implications}

\subsection{Does Scale Matter?}
\label{sec:scale_matter}

We train BART\textsubscript{BASE} using the same task decomposition and knowledge infusion methods reported earlier. We find that the results do not fully transfer to smaller models. Specifically, the results are only better when the training examples are 16 to 64, otherwise the baseline model surpasses the \token model. Showing that scale does matter. The results are shown in Table \ref{tab:scale_results} in the Appendix ($\S$\ref{app:scale}).

\begin{figure*}
    \centering
    \includegraphics[width=1\linewidth]{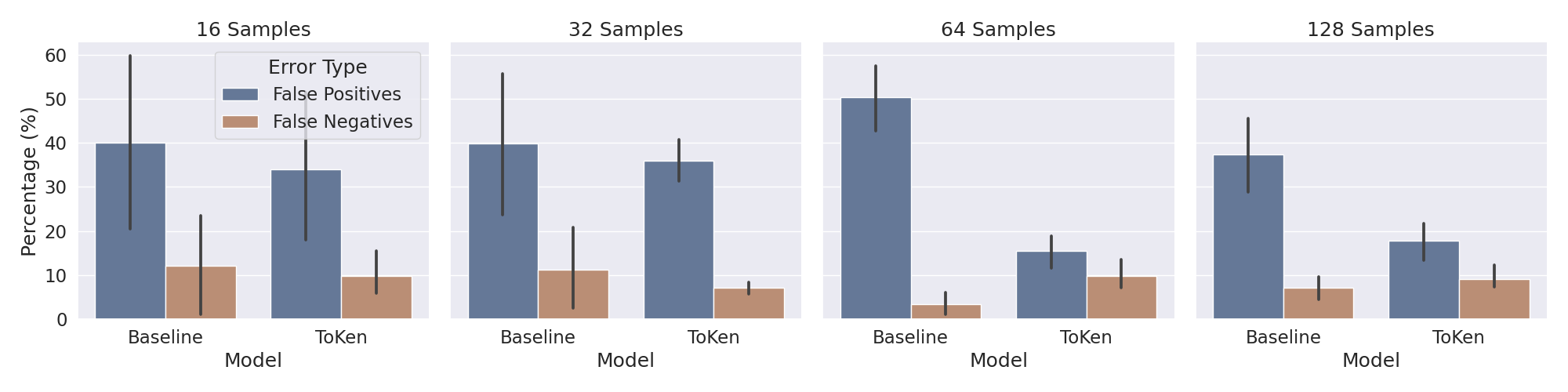}
    \caption{The percentage of \textit{False Positives} and \textit{False Negatives} across 10 runs on the testing set.}
    \label{fig:error_analysis}
\end{figure*}

\subsection{Error Analysis}

Figure \ref{fig:error_analysis} shows the percentage of false positives (FPs) and false negatives (FNs) across 10 runs on the testing set for the 16 to 128-shot case. It can be seen that the difference in performance between both models is largely due to a considerable decrease in the percentage of FPs for the \token model as we increase the number of samples, while the percentage of FNs stay almost constant. However, the percentage of FNs are always less than the FPs, which is a desirable property for HS models as FPs can be more tolerated (i.e. over enforcement) than FNs (i.e. under enforcement), since the latter will lead to more propagation of hateful content. As discussed in Section \ref{sec:perf_analysis}, predicting the group identity in the \token model enables the model to correctly detect whether the post is targeting a group or an individual, and we can see that clearly in posts where the baseline model got a FN and the \token model got the correct result. For example, the baseline model wasn't able to predict this post as HS: ``\textit{Your approval is so worthless you should pay people to take it @user Call me sexist but I do not approve of female football presenters}'', while the subtasks model identified correctly that it targets \code{Women} leading to a correct prediction. On the other hand, this is an example that the baseline predicted correctly while the \token model did not: ``\textit{What do you get when an epileptic falls into a garden? A seizure salad.}'' The reason for this is the \token model predicted that this post targets an \code{Individual} and not a \code{Group}.

\section{Related Work}
Social media provides a platform for users to connect with people all over the world and engage in ways that were not previously possible. Recent surveys show that $41\%$ of internet users experienced some form of harassment online, with a third of these cases being identity-related (i.e. race, gender, sexual orientation, etc.) \cite{vogels2021state,league2020online}. The sheer scale of content shared on social media platforms makes manual moderation untenable and necessitates automated methods for detecting hateful content \cite{halevy2022preserving}. This has led to an increased interest in automated hate speech detection, both in terms of collecting corpora \cite{poletto2021resources} as well as improved methods for hate speech detection \cite{schmidt-wiegand-2017-survey,fortuna2018survey}.


Early work in hate speech detection has treated the problem as a binary classification task, requiring annotators to simply indicate whether or not a given post constitutes hate speech \cite{waseem-hovy-2016-hateful, Davidson2017AutomatedHS,founta2018large}. However, recent work has shown that eliciting binary judgments for hate speech is unreliable and leads to poor inter-annotator agreement \cite{sanguinetti-etal-2018-italian,assimakopoulos2020annotating}. This has lead to increased work in collecting hate speech annotation with more complex annotation schemas. \newcite{zampieri-etal-2019-predicting} propose a three-level annotation schema that identifies both the type and target of offensiveness in social media posts. Another line of work proposes a hierarchical annotation schema where the task of determining hate speech is broken down into subtasks, in an effort to eliminate some of the subjectivity \cite{assimakopoulos2020annotating, sap2020socialbiasframes}. \newcite{rahman2021information} combine established information retrieval techniques with task decomposition and annotator rationale, in order to create a higher quality dataset for hate speech detection. While the aforementioned studies explore the idea of task decomposition in improving annotation consistency, our work instead looks at the role of task decomposition in building more robust, generalizable models for few-shot hate speech detection.


The focus of the limited prior work on few-shot hate speech detection has been to explore zero-shot/few-shot crosslingual transfer from a source language (such as English) with sufficient hate speech data to a target language with limited data \cite{Stappen2020CrosslingualZA,nozza-2021-exposing}. In contrast, our work explores how task decomposition and knowledge infusion can help even when there is not sufficient hate speech data in English.

\section{Conclusion}

In this work, we propose \token, a method to train language models for detecting HS in the few-shot setting. We show that it significantly outperforms comparable baseline models that predicts the HS label directly instead of decomposing it into its constituent parts. We further show that task decomposition not only improves the performance, but also allows for fine-grained inspection of the model's behavior. Since HS is a complex phenomenon that requires a set of reasoning skills not readily available in such pre-trained models, we pre-finetune the BART\textsubscript{LARGE} on both the \atomic and StereoSet datasets to equip the model with commonsense reasoning and knowledge of stereotypes that we show leads to further improvement in HS detection performance. We show that the \token models generalize better to three out-of-distribution datasets in the few-shot setting as well as being significantly more robust to training setups. We further analyze the model's behavior in terms of the order in which the HS labels appears, the scale of the model and the performance when adding an additional subtask that explains the implied stereotype of the post. 

In future work, we plan on investigating the role of task decomposition, knowledge infusion and the additional subtask of explaining the implication behind the post in large language models as well as explore the \token method in low-resource languages, where it is expected to be most beneficial.

\section{Limitations}

We note a few limitations of our work: (1) in our experiments we compared our task-decomposed model to standard models as baselines. It will be valuable for future work to compare our models against other models of similar scale trained using multi-task learning in a similar manner to \cite{AlKhamissi2022MetaAA}, where each classification head is subtask-specific and trained using categorical cross-entropy on the corresponding number of classes. However, that would require categorizing the group identities into a discrete number of classes. (2) To the best of our knowledge there is no literature that uses the SBIC dataset in a few-shot hate speech setting, therefore we resorted to the baseline with binary prediction using the same conditional generation framework. Future work should compare with such models. (3) The datasets used in this work are mostly looking at HS from a western perspective and are only in English. Different languages and societies may have subtleties which may affect the performance of HS systems. Even though we believe that our work is generalizable beyond the English language, we have not evaluated this, and we encourage future work to look beyond the settings we explored in this paper. (4) We follow prior work and determine hate-speech labels based on majority vote which might silence the voice of minority groups, which is especially problematic in this context. In the future we hope to model dissenting opinions between the annotators similar to recent work \cite{gordon2022jury}. 

\bibliography{anthology}
\bibliographystyle{acl_natbib}

\appendix
\section{Appendix}

\subsection{\atomic Human Readable Templates}
\label{app:atomic_templates}

The \atomic knowledge graph consists of tuples in the form of $\langle$ \texttt{Head}, \texttt{Relation}, \texttt{Tail} $\rangle$, where \texttt{Head} and \texttt{Tail} are provided in textual format. However, we found that training BART on this dataset directly led to sub-optimal results; where it is tasked to predict the \texttt{Tail} given the \texttt{Head} and \texttt{Relation}. Instead we replaced each relation with a human readable template as shown in Table \ref{tab:atomic_templates}, where the \texttt{Head} is injected in place of \{0\} and the model is tasked with predicting the \texttt{Tail}.
\input{tables/atomic_templates}

\input{tables/scale_table}

\subsection{StereoSet Training}
\label{app:stereoset}

\citet{Nadeem2021StereoSetMS} designed two types of \textit{Context Association Tests} (CATs): intrasentence and intersentence to measure the degree of stereotypical bias in language models at the sentence level and discourse level respectively. In this work, we only consider the latter and only include stereotypical sentences (i.e. omit the anti-stereotype examples). In the pre-finetuning step of StereoSet, the BART model is tasked with predicting the explanation of a given stereotype conditioned on the targeted group, bias type and stereotype. The dataset is linearized according to the following format: \textbf{Input}: "Target: \texttt{\{TARGET\_GROUP\}}  Bias Type: \texttt{\{BIAS\_TYPE\}} Sentence: \texttt{\{STEREOTYPE\}} Context?" and the \textbf{Output}: is "\texttt{\{EXPLANATION\}}". For example: \textbf{Input}: "Target: \{Cape Verde\} Bias Type: \{race\} Sentence: \{They rely on the ports.\} Context?"; \textbf{Output}: "\{Cape Verde has a lot of water.\}".

\subsection{Scale Experiments}
\label{app:scale}
In this analysis, we investigate the role of scale in \token. Table \ref{tab:scale_results} shows that the results are only better from the 16 to the 64-shot case, suggesting that scale does play a role in the proposed method.

\end{document}

%% file: tables/task_decomposition.tex
\begin{table*}[ht]
\centering
\begin{tabular}{@{}lccccccc@{}}
\toprule
\textbf{Model}               & \textbf{16}      & \textbf{32}      & \textbf{64}      & \textbf{128}     & \textbf{256}     & \textbf{512}     & \textbf{1024}    \\ \midrule
\textbf{Baseline}           & 45.31          & 53.23          & 56.41          & 60.12          & 64.37          & 70.29          & \textbf{73.95} \\
\textbf{Minimal Decomposition}  & 50.79          & 56.12          & 56.46          & 59.78          & 64.94          & 67.83          & 69.95          \\
\textbf{+ Group Identification} & \textbf{58.89} & \textbf{61.77} & \textbf{68.03} & \textbf{70.25} & \textbf{70.28} & \textbf{70.65} & 72.76          \\ \bottomrule
\end{tabular}
\caption{Results of the Task Decomposed Model on the \textit{Hate Speech} detection task (Binary F1-score). \textbf{Baseline} predicts only whether the input post is HS or not. \textbf{Minimal Decomposition} additionally predicts whether the post is offensive or not and the group detection. \textbf{+ Group Identification} additionally predicts the minority groups the post is targeting if any. }
\label{tab:task_decomposed_results}
\end{table*}

%% file: tables/knowledge_infusion.tex
\begin{table*}
\centering
\begin{tabular}{@{}llccccccc@{}}
\toprule
\multicolumn{2}{c}{\textbf{Model}}                        & \textbf{16}                 & \textbf{32}                 & \textbf{64}                          & \textbf{128}                         & \textbf{256}                         & \textbf{512}                & \textbf{1024}               \\ \midrule
\multirow{2}{*}{\textbf{BART}}        & \textbf{Baseline} & 45.31                      & 53.23                     & 56.41                             & 60.12                                & 64.37                            & 70.29                    & 73.95           \\
                                      & \textbf{Subtasks} & 58.89                      & 61.77                     & 68.03                            &  70.25                                & 70.28                            & 70.65          & 72.76                      \\ \midrule \midrule
\multirow{2}{*}{\textbf{+ StereoSet}} & \textbf{Baseline} & 53.30                      & 54.68                     & 54.17                              & 61.41                              & 67.69                              & 71.25            & 73.68                     \\
                                      & \textbf{Subtasks} & 42.86                      & \textbf{66.17}            & 69.01                              & 70.06                              & 70.14                              & 72.14                     & 72.64                     \\ \midrule \midrule
\multirow{2}{*}{\textbf{+ \atomic}}     & \textbf{Baseline} & 44.76                      & 49.60                      & 64.89                              & 69.38                              & 70.09                              & 72.32            & \textbf{73.97}                     \\
                                      & \textbf{Subtasks} & \textbf{63.14}             & 62.01                    & 67.96                              & 70.94                              & 70.16                              & 72.29                     & 72.96                     \\ 
\multirow{2}{*}{\textbf{+ StereoSet}} & \textbf{Baseline} & 44.75 & 47.47 & 56.18          & 62.88          & 66.38          & 69.77 & 71.50 \\
                                      & \textbf{Subtasks} & 59.74 & 63.28 & \textbf{70.08} & \textbf{70.99} & \textbf{70.57} & \textbf{72.36} & 73.80 \\ \bottomrule
\end{tabular}
\caption{\textbf{Knowledge Infusion Results} Here we report binary F1-score on the SBIC testing set for the HS detection task using models with different degrees of knowledge infusion. In each row we compare the corresponding baseline and subtasks models. Results in \textbf{bold} show the best overall model in each few-shot setting. See Section \ref{sec:knowledge_infusion} for more details.}
\label{tab:knowledge_infusion_results}
\end{table*}

%% file: tables/hate_speech_position.tex
\begin{table*}[ht]
\centering
\begin{tabular}{@{}lccccccc@{}}
\toprule
\textbf{Order}               & \textbf{16}      & \textbf{32}      & \textbf{64}      & \textbf{128}     & \textbf{256}     & \textbf{512}     & \textbf{1024}    \\ \midrule
OFF GD GI \textbf{HS}           & \textbf{55.60}         & \textbf{62.31}          & \textbf{68.47}          & 69.22          & 69.64          & 70.49          & 71.69 \\
OFF GD \textbf{HS} GI  & 54.67          & 60.36          & 68.02          & 67.65          & 68.66          & 70.13          & 70.92          \\
OFF \textbf{HS} GD GI & 54.53         & 56.14            & 67.02          & 68.59           & \textbf{71.37}               & \textbf{72.47} & \textbf{72.39}          \\ 
\textbf{HS} OFF GD GI & 51.64         & 62.04           & 64.48           & \textbf{69.45}           & 70.33            & 71.33   & 72.20          \\ \midrule

GD GI \textbf{OFF} HS & 38.28          & 27.11          & 31.32          & 53.98          & 52.12          & 60.69          & 67.20   \\ \bottomrule
\end{tabular}
\caption{The validation performance of the best model on the HS detection task as a function of the position of the HS label in the sequence of subtasks across different number of training samples.}
\label{tab:hs_pos_results}
\end{table*}

%% file: tables/implications.tex
\begin{table}
\centering
\begin{tabular}{@{}cccc@{}}
\toprule
\textbf{\# of Samples} & \textbf{Baseline} & \textbf{\token} & \textbf{+Impl} \\ \midrule
\textbf{16}            & 52.67           & 58.21           & 57.02                \\
\textbf{32}            & 52.71           & 64.47           & 59.98                \\
\textbf{64}            & 57.60           & 70.93           & 65.50                \\
\textbf{128}           & 60.01           & 71.25           & 67.89                \\
\textbf{256}           & 66.81           & 71.62           & 69.22                \\
\textbf{512}           & 69.41           & 72.59           & 70.12                \\
\textbf{1024}          & 74.72           & 74.09           & 71.45                \\ \bottomrule
\end{tabular}
\caption{\textbf{Implications Results} The HS detection performance of the \textbf{Baseline} in comparison with the \textbf{Subtasks} models before and after adding the implication to the subtasks across 5 runs for a given training set size.}
\label{tab:implications}
\end{table}

%% file: tables/atomic_templates.tex
\begin{table}
\begin{tabular}{@{}p{0.28\linewidth}p{0.65\linewidth}@{}}
\toprule
\textbf{Relation} & \textbf{Human Readable Template}                  \\ \midrule
ObjectUse          & \{0\} is used for \{1\}                           \\
AtLocation         & You are likely to find \{0\} in \{1\}             \\
MadeUpOf           & \{0\} is made up of \{1\}                         \\
HasProperty        & \{0\} is \{1\}                                    \\
CapableOf          & \{0\} can \{1\}                                   \\
Desires            & \{0\} wants \{1\}                                 \\
NotDesires         & \{0\} does not want \{1\}                         \\
isAfter            & Something that happens after \{0\} is \{1\}       \\
HasSubEvent        & Something you might do while \{0\} is \{1\}       \\
isBefore           & Something that happens before \{0\} is \{1\}      \\
HinderedBy         & \{0\} is hindered by \{1\}                        \\
Causes             & Sometimes \{0\} causes \{1\}                      \\
xReason            & \{0\}. The reason for PersonX doing this is \{1\} \\
isFilledBy         & \{0\} can be filled by \{1\}                      \\
xNeed              & But before \{0\}, PersonX needed \{1\}            \\
xAttr              & \{0\} is seen as \{1\}                            \\
xEffect            & As a result of \{0\}, PersonX will \{1\}          \\
xReact             & As a result of \{0\}, PersonX feels \{1\}         \\
xWant              & After \{0\}, PersonX would want \{1\}             \\
xIntent            & Because of \{0\}, PersonX wanted \{1\}            \\
oEffect            & as a result of \{0\}, others will \{1\}           \\
oReact             & as a result of \{0\}, others would feel \{1\}     \\
oWant              & as a result of \{0\}, others would want \{1\}     \\ \bottomrule
\end{tabular}
\caption{Human readable templates for each relation used to train the BART model. The \texttt{Head} is injected in place of \{0\} and is tasked with predicting the \texttt{Tail} \{1\} in a conditional generation framework.}
\label{tab:atomic_templates}
\end{table}

%% file: tables/scale_table.tex
\begin{table*}[ht]
\centering
\begin{tabular}{@{}lccccccc@{}}
\toprule
\textbf{Model}               & \textbf{16}      & \textbf{32}      & \textbf{64}      & \textbf{128}     & \textbf{256}     & \textbf{512}     & \textbf{1024}    \\ \midrule
\textbf{Baseline}           & 53.49          & 51.47          & 58.28          & \textbf{66.70}         & \textbf{68.24}          & \textbf{70.09}          & \textbf{71.30} \\
\textbf{\token}           & \textbf{55.23}       & \textbf{60.29}         & \textbf{63.68}         & 65.10          & 67.52          & 68.67          & 69.66          \\
\end{tabular}
\caption{Results using BART\textsubscript{BASE} as the core model. Similar to the previous experiments the baseline model predicts the HS label directly, while the \token model employs task decomposition and knowledge infusion using the \atomic and StereoSet datasets. (see $\S$\ref{sec:scale_matter} for more details)}
\label{tab:scale_results}
\end{table*}

%% file: paper.bbl
\begin{thebibliography}{31}
\expandafter\ifx\csname natexlab\endcsname\relax\def\natexlab#1{#1}\fi

\bibitem[{AlKhamissi and Diab(2022)}]{AlKhamissi2022MetaAA}
Badr AlKhamissi and Mona~T. Diab. 2022.
\newblock Meta ai at arabic hate speech 2022: Multitask learning with
  self-correction for hate speech classification.
\newblock \emph{ArXiv}, abs/2205.07960.

\bibitem[{Assimakopoulos et~al.(2020{\natexlab{a}})Assimakopoulos, Muskat,
  van~der Plas, and Gatt}]{assimakopoulos2020annotating}
Stavros Assimakopoulos, Rebecca~Vella Muskat, Lonneke van~der Plas, and Albert
  Gatt. 2020{\natexlab{a}}.
\newblock Annotating for hate speech: The maneco corpus and some input from
  critical discourse analysis.
\newblock \emph{arXiv preprint arXiv:2008.06222}.

\bibitem[{Assimakopoulos et~al.(2020{\natexlab{b}})Assimakopoulos,
  Vella~Muskat, van~der Plas, and Gatt}]{assimakopoulos-etal-2020-annotating}
Stavros Assimakopoulos, Rebecca Vella~Muskat, Lonneke van~der Plas, and Albert
  Gatt. 2020{\natexlab{b}}.
\newblock \href {https://aclanthology.org/2020.lrec-1.626} {Annotating for hate
  speech: The {M}a{N}e{C}o corpus and some input from critical discourse
  analysis}.
\newblock In \emph{Proceedings of the 12th Language Resources and Evaluation
  Conference}, pages 5088--5097, Marseille, France. European Language Resources
  Association.

\bibitem[{Attanasio et~al.(2022)Attanasio, Nozza, Pastor, and
  Hovy}]{attanasio-etal-2022-benchmarking}
Giuseppe Attanasio, Debora Nozza, Eliana Pastor, and Dirk Hovy. 2022.
\newblock \href {https://doi.org/10.18653/v1/2022.nlppower-1.11} {Benchmarking
  post-hoc interpretability approaches for transformer-based misogyny
  detection}.
\newblock In \emph{Proceedings of NLP Power! The First Workshop on Efficient
  Benchmarking in NLP}, pages 100--112, Dublin, Ireland. Association for
  Computational Linguistics.

\bibitem[{Davidson et~al.(2019)Davidson, Bhattacharya, and
  Weber}]{davidson-etal-2019-racial}
Thomas Davidson, Debasmita Bhattacharya, and Ingmar Weber. 2019.
\newblock \href {https://doi.org/10.18653/v1/W19-3504} {Racial bias in hate
  speech and abusive language detection datasets}.
\newblock In \emph{Proceedings of the Third Workshop on Abusive Language
  Online}, pages 25--35, Florence, Italy. Association for Computational
  Linguistics.

\bibitem[{Davidson et~al.(2017)Davidson, Warmsley, Macy, and
  Weber}]{Davidson2017AutomatedHS}
Thomas Davidson, Dana Warmsley, Michael~W. Macy, and Ingmar Weber. 2017.
\newblock Automated hate speech detection and the problem of offensive
  language.
\newblock In \emph{ICWSM}.

\bibitem[{de~Gibert et~al.(2018)de~Gibert, Perez, Garc{\'\i}a-Pablos, and
  Cuadros}]{gibert2018hate}
Ona de~Gibert, Naiara Perez, Aitor Garc{\'\i}a-Pablos, and Montse Cuadros.
  2018.
\newblock \href {https://doi.org/10.18653/v1/W18-5102} {{Hate Speech Dataset
  from a White Supremacy Forum}}.
\newblock In \emph{Proceedings of the 2nd Workshop on Abusive Language Online
  ({ALW}2)}, pages 11--20, Brussels, Belgium. Association for Computational
  Linguistics.

\bibitem[{ElSherief et~al.(2021)ElSherief, Ziems, Muchlinski, Anupindi,
  Seybolt, De~Choudhury, and Yang}]{elsherief-etal-2021-latent}
Mai ElSherief, Caleb Ziems, David Muchlinski, Vaishnavi Anupindi, Jordyn
  Seybolt, Munmun De~Choudhury, and Diyi Yang. 2021.
\newblock \href {https://doi.org/10.18653/v1/2021.emnlp-main.29} {Latent
  hatred: A benchmark for understanding implicit hate speech}.
\newblock In \emph{Proceedings of the 2021 Conference on Empirical Methods in
  Natural Language Processing}, pages 345--363, Online and Punta Cana,
  Dominican Republic. Association for Computational Linguistics.

\bibitem[{Fortuna and Nunes(2018)}]{fortuna2018survey}
Paula Fortuna and S{\'e}rgio Nunes. 2018.
\newblock A survey on automatic detection of hate speech in text.
\newblock \emph{ACM Computing Surveys (CSUR)}, 51(4):1--30.

\bibitem[{Founta et~al.(2018)Founta, Djouvas, Chatzakou, Leontiadis, Blackburn,
  Stringhini, Vakali, Sirivianos, and Kourtellis}]{founta2018large}
Antigoni~Maria Founta, Constantinos Djouvas, Despoina Chatzakou, Ilias
  Leontiadis, Jeremy Blackburn, Gianluca Stringhini, Athena Vakali, Michael
  Sirivianos, and Nicolas Kourtellis. 2018.
\newblock Large scale crowdsourcing and characterization of twitter abusive
  behavior.
\newblock In \emph{Twelfth International AAAI Conference on Web and Social
  Media}.

\bibitem[{Gordon et~al.(2022)Gordon, Lam, Park, Patel, Hancock, Hashimoto, and
  Bernstein}]{gordon2022jury}
Mitchell~L Gordon, Michelle~S Lam, Joon~Sung Park, Kayur Patel, Jeff Hancock,
  Tatsunori Hashimoto, and Michael~S Bernstein. 2022.
\newblock Jury learning: Integrating dissenting voices into machine learning
  models.
\newblock In \emph{CHI Conference on Human Factors in Computing Systems}, pages
  1--19.

\bibitem[{Halevy et~al.(2022)Halevy, Canton-Ferrer, Ma, Ozertem, Pantel,
  Saeidi, Silvestri, and Stoyanov}]{halevy2022preserving}
Alon Halevy, Cristian Canton-Ferrer, Hao Ma, Umut Ozertem, Patrick Pantel,
  Marzieh Saeidi, Fabrizio Silvestri, and Ves Stoyanov. 2022.
\newblock Preserving integrity in online social networks.
\newblock \emph{Communications of the ACM}, 65(2):92--98.

\bibitem[{Hwang et~al.(2021)Hwang, Bhagavatula, Bras, Da, Sakaguchi, Bosselut,
  and Choi}]{Hwang2021COMETATOMIC2O}
Jena~D. Hwang, Chandra Bhagavatula, Ronan~Le Bras, Jeff Da, Keisuke Sakaguchi,
  Antoine Bosselut, and Yejin Choi. 2021.
\newblock Comet-atomic 2020: On symbolic and neural commonsense knowledge
  graphs.
\newblock In \emph{AAAI}.

\bibitem[{League(2020)}]{league2020online}
Anti-Defamation League. 2020.
\newblock Online hate and harassment. the american experience 2021.
\newblock \emph{Center for Technology and Society. Retrieved from www. adl.
  org/media/14643/download}.

\bibitem[{Lewis et~al.(2020)Lewis, Liu, Goyal, Ghazvininejad, Mohamed, Levy,
  Stoyanov, and Zettlemoyer}]{lewis-etal-2020-bart}
Mike Lewis, Yinhan Liu, Naman Goyal, Marjan Ghazvininejad, Abdelrahman Mohamed,
  Omer Levy, Veselin Stoyanov, and Luke Zettlemoyer. 2020.
\newblock \href {https://doi.org/10.18653/v1/2020.acl-main.703} {{BART}:
  Denoising sequence-to-sequence pre-training for natural language generation,
  translation, and comprehension}.
\newblock In \emph{Proceedings of the 58th Annual Meeting of the Association
  for Computational Linguistics}, pages 7871--7880, Online. Association for
  Computational Linguistics.

\bibitem[{Mathew et~al.(2021)Mathew, Saha, Yimam, Biemann, Goyal, and
  Mukherjee}]{mathew2020hatexplain}
Binny Mathew, Punyajoy Saha, Seid~Muhie Yimam, Chris Biemann, Pawan Goyal, and
  Animesh Mukherjee. 2021.
\newblock Hatexplain: A benchmark dataset for explainable hate speech
  detection.
\newblock In \emph{AAAI conference on artificial intelligence}.

\bibitem[{Mollas et~al.(2020{\natexlab{a}})Mollas, Chrysopoulou, Karlos, and
  Tsoumakas}]{Mollas2020ETHOSAO}
Ioannis Mollas, Zoe Chrysopoulou, Stamatis Karlos, and Grigorios Tsoumakas.
  2020{\natexlab{a}}.
\newblock Ethos: an online hate speech detection dataset.
\newblock \emph{ArXiv}, abs/2006.08328.

\bibitem[{Mollas et~al.(2020{\natexlab{b}})Mollas, Chrysopoulou, Karlos, and
  Tsoumakas}]{mollas2020ethos}
Ioannis Mollas, Zoe Chrysopoulou, Stamatis Karlos, and Grigorios Tsoumakas.
  2020{\natexlab{b}}.
\newblock Ethos: an online hate speech detection dataset.
\newblock \emph{arXiv preprint arXiv:2006.08328}.

\bibitem[{Nadeem et~al.(2021)Nadeem, Bethke, and Reddy}]{Nadeem2021StereoSetMS}
Moin Nadeem, Anna Bethke, and Siva Reddy. 2021.
\newblock Stereoset: Measuring stereotypical bias in pretrained language
  models.
\newblock In \emph{ACL/IJCNLP}.

\bibitem[{Nozza(2021)}]{nozza-2021-exposing}
Debora Nozza. 2021.
\newblock \href {https://doi.org/10.18653/v1/2021.acl-short.114} {Exposing the
  limits of zero-shot cross-lingual hate speech detection}.
\newblock In \emph{Proceedings of the 59th Annual Meeting of the Association
  for Computational Linguistics and the 11th International Joint Conference on
  Natural Language Processing (Volume 2: Short Papers)}, pages 907--914,
  Online. Association for Computational Linguistics.

\bibitem[{Poletto et~al.(2021)Poletto, Basile, Sanguinetti, Bosco, and
  Patti}]{poletto2021resources}
Fabio Poletto, Valerio Basile, Manuela Sanguinetti, Cristina Bosco, and Viviana
  Patti. 2021.
\newblock Resources and benchmark corpora for hate speech detection: a
  systematic review.
\newblock \emph{Language Resources and Evaluation}, 55(2):477--523.

\bibitem[{Rahman et~al.(2021)Rahman, Balakrishnan, Murthy, Kutlu, and
  Lease}]{rahman2021information}
Md~Mustafizur Rahman, Dinesh Balakrishnan, Dhiraj Murthy, Mucahid Kutlu, and
  Matthew Lease. 2021.
\newblock An information retrieval approach to building datasets for hate
  speech detection.
\newblock \emph{arXiv preprint arXiv:2106.09775}.

\bibitem[{Sanguinetti et~al.(2018)Sanguinetti, Poletto, Bosco, Patti, and
  Stranisci}]{sanguinetti-etal-2018-italian}
Manuela Sanguinetti, Fabio Poletto, Cristina Bosco, Viviana Patti, and Marco
  Stranisci. 2018.
\newblock \href {https://aclanthology.org/L18-1443} {An {I}talian {T}witter
  corpus of hate speech against immigrants}.
\newblock In \emph{Proceedings of the Eleventh International Conference on
  Language Resources and Evaluation ({LREC} 2018)}, Miyazaki, Japan. European
  Language Resources Association (ELRA).

\bibitem[{Sap et~al.(2019)Sap, Card, Gabriel, Choi, and
  Smith}]{sap-etal-2019-risk}
Maarten Sap, Dallas Card, Saadia Gabriel, Yejin Choi, and Noah~A. Smith. 2019.
\newblock \href {https://doi.org/10.18653/v1/P19-1163} {The risk of racial bias
  in hate speech detection}.
\newblock In \emph{Proceedings of the 57th Annual Meeting of the Association
  for Computational Linguistics}, pages 1668--1678, Florence, Italy.
  Association for Computational Linguistics.

\bibitem[{Sap et~al.(2020)Sap, Gabriel, Qin, Jurafsky, Smith, and
  Choi}]{sap2020socialbiasframes}
Maarten Sap, Saadia Gabriel, Lianhui Qin, Dan Jurafsky, Noah~A Smith, and Yejin
  Choi. 2020.
\newblock Social bias frames: Reasoning about social and power implications of
  language.
\newblock In \emph{ACL}.

\bibitem[{Schmidt and Wiegand(2017)}]{schmidt-wiegand-2017-survey}
Anna Schmidt and Michael Wiegand. 2017.
\newblock \href {https://doi.org/10.18653/v1/W17-1101} {A survey on hate speech
  detection using natural language processing}.
\newblock In \emph{Proceedings of the Fifth International Workshop on Natural
  Language Processing for Social Media}, pages 1--10, Valencia, Spain.
  Association for Computational Linguistics.

\bibitem[{Stappen et~al.(2020)Stappen, Brunn, and
  Schuller}]{Stappen2020CrosslingualZA}
Lukas Stappen, Fabian Brunn, and Bj{\"o}rn~W. Schuller. 2020.
\newblock Cross-lingual zero- and few-shot hate speech detection utilising
  frozen transformer language models and axel.
\newblock \emph{ArXiv}, abs/2004.13850.

\bibitem[{Vogels(2021)}]{vogels2021state}
Emily~A Vogels. 2021.
\newblock The state of online harassment.
\newblock \emph{Pew Research Center}, 13.

\bibitem[{Waseem and Hovy(2016)}]{waseem-hovy-2016-hateful}
Zeerak Waseem and Dirk Hovy. 2016.
\newblock \href {https://doi.org/10.18653/v1/N16-2013} {Hateful symbols or
  hateful people? predictive features for hate speech detection on {T}witter}.
\newblock In \emph{Proceedings of the {NAACL} Student Research Workshop}, pages
  88--93, San Diego, California. Association for Computational Linguistics.

\bibitem[{Wolf et~al.(2019)Wolf, Debut, Sanh, Chaumond, Delangue, Moi, Cistac,
  Rault, Louf, Funtowicz, and Brew}]{Wolf2019HuggingFace}
Thomas Wolf, Lysandre Debut, Victor Sanh, Julien Chaumond, Clement Delangue,
  Anthony Moi, Pierric Cistac, Tim Rault, R'emi Louf, Morgan Funtowicz, and
  Jamie Brew. 2019.
\newblock \href {http://arxiv.org/abs/1910.03771} {Huggingface's transformers:
  State-of-the-art natural language processing}.
\newblock \emph{ArXiv}, abs/1910.03771.

\bibitem[{Zampieri et~al.(2019)Zampieri, Malmasi, Nakov, Rosenthal, Farra, and
  Kumar}]{zampieri-etal-2019-predicting}
Marcos Zampieri, Shervin Malmasi, Preslav Nakov, Sara Rosenthal, Noura Farra,
  and Ritesh Kumar. 2019.
\newblock \href {https://doi.org/10.18653/v1/N19-1144} {Predicting the type and
  target of offensive posts in social media}.
\newblock In \emph{Proceedings of the 2019 Conference of the North {A}merican
  Chapter of the Association for Computational Linguistics: Human Language
  Technologies, Volume 1 (Long and Short Papers)}, pages 1415--1420,
  Minneapolis, Minnesota. Association for Computational Linguistics.

\end{thebibliography}
